\documentclass[sigconf]{acmart}

\AtBeginDocument{%
  }

\copyrightyear{2023}
\acmYear{2023}
\setcopyright{rightsretained}

\acmConference[MMAsia '23]{ACM Multimedia Asia 2023}{December 6--8,
2023}{Tainan, Taiwan}
\acmBooktitle{ACM Multimedia Asia 2023 (MMAsia '23), December 6--8, 2023,
Tainan, Taiwan}
\acmDOI{10.1145/3595916.3626371}
\acmISBN{979-8-4007-0205-1/23/12}

\usepackage{bm}
\usepackage{graphicx}
\usepackage{amsmath}

\usepackage{amssymb}
\usepackage{amsfonts}
\usepackage{algorithm}  
\usepackage{booktabs}

\usepackage[utf8]{inputenc} % allow utf-8 input
\usepackage[T1]{fontenc}    % use 8-bit T1 fonts
\usepackage{hyperref}       % hyperlinks
\usepackage{url}            % simple URL typesetting
\usepackage{booktabs}       % professional-quality tables
\usepackage{amsfonts}       % blackboard math symbols
\usepackage{nicefrac}       % compact symbols for 1/2, etc.
\usepackage{microtype}      % microtypography
\usepackage{xcolor}         % colors
\usepackage{subcaption}

\usepackage[noend]{algpseudocode}
\usepackage{multirow}

\newcommand{\myPara}[1]{\noindent\textbf{#1}}

\graphicspath{{./fig/}}

\begin{document}

%%
%% The "title" command has an optional parameter,
%% allowing the author to define a "short title" to be used in page headers.
\title{Personalized Federated Learning via Backbone Self-Distillation}

%%
%% The "author" command and its associated commands are used to define
%% the authors and their affiliations.
%% Of note is the shared affiliation of the first two authors, and the
%% "authornote" and "authornotemark" commands
%% used to denote shared contribution to the research.
\author{Pengju Wang}
\affiliation{%
  \institution{Institute of Information Engineering, Chinese Academy of Sciences \\ School of Cyber Security, UCAS}
  \country{}
  \city{}
}
\email{wangpengju@iie.ac.cn}

\author{Bochao Liu}
\affiliation{%
  \institution{Institute of Information Engineering, Chinese Academy of Sciences \\ School of Cyber Security, UCAS}
  \country{}
  \city{}
}
\email{liubochao@iie.ac.cn}

\author{Dan Zeng}
\affiliation{%
  \institution{School of Communication and Information Engineering, Shanghai University}
  \country{}
  \city{}
}
\email{dzeng@shu.edu.cn}

\author{Chenggang Yan}
\affiliation{%
  \institution{School of Automation, Hangzhou Dianzi University }
  \country{}
  \city{}
}
\email{cgyan@hdu.edu.cn}

\author{Shiming Ge}\authornote{Shiming Ge is the corresponding author (geshiming@iie.ac.cn).}
\affiliation{%
  \institution{Institute of Information Engineering, Chinese Academy of Sciences \\ School of Cyber Security, UCAS}
  \country{}
  \city{}
}
\email{geshiming@iie.ac.cn}

\renewcommand{\shortauthors}{Pengju Wang et al.}

%%
%% By default, the full list of authors will be used in the page
%% headers. Often, this list is too long, and will overlap
%% other information printed in the page headers. This command allows
%% the author to define a more concise list
%% of authors' names for this purpose.
% \renewcommand{\shortauthors}{Trovato et al.}

%%
%% The abstract is a short summary of the work to be presented in the
%% article.
\begin{abstract}
In practical scenarios, federated learning frequently necessitates training personalized models for each client using heterogeneous data. This paper proposes a backbone self-distillation approach to facilitate personalized federated learning. In this approach, each client trains its local model and only sends the backbone weights to the server. These weights are then aggregated to create a global backbone, which is returned to each client for updating. However, the client's local backbone lacks personalization because of the common representation. To solve this problem, each client further performs backbone self-distillation by using the global backbone as a teacher and transferring knowledge to update the local backbone. This process involves learning two components: the shared backbone for common representation and the private head for local personalization, which enables effective global knowledge transfer. Extensive experiments and comparisons with 12 state-of-the-art approaches demonstrate the effectiveness of our approach.
\end{abstract}

%%
%% The code below is generated by the tool at http://dl.acm.org/ccs.cfm.
%% Please copy and paste the code instead of the example below.
%%


%%
%% Keywords. The author(s) should pick words that accurately describe
%% the work being presented. Separate the keywords with commas.
%% A "teaser" image appears between the author and the affiliation
%% information and the body of the document, and typically spans the
%% page.

% \received{20 February 2007}
% \received[revised]{12 March 2009}
% \received[accepted]{5 June 2009}

%%
%% This command processes the author and affiliation and title
%% information and builds the first part of the formatted document.
\maketitle

\section{Introduction}
Deep learning models have achieved remarkable performance in numerous significant tasks~\cite{pouyanfar2018survey}, primarily due to the availability of large-scale data. Nevertheless, in many practical scenarios like multimedia analysis~\cite{zhuang2020performance}, medical diagnosis~\cite{karargyris2023federated}, and financial analytics~\cite{chatterjee2023use}, access to multimedia data is limited for model training because of privacy issues~\cite{li2020invisiblefl, qi2022feeling}. Federated learning~\cite{mcmahan2017communication} has recently been proposed as a practical and feasible solution for collaborative training of models on distributed data from multiple clients while preserving privacy. The core concept revolves around sharing knowledge, such as weights, as opposed to sharing data among multiple clients through a centralized server. However, an important challenge facing current federated learning is the data heterogeneity problem~\cite{kairouz2019advances}. It is mainly manifested as different data distributions due to different local tasks of clients. Traditional federated learning only learns a common global model, but non-independent and identically distributed data would lead to poor performance for each client, i.e., client drift~\cite{karimireddy2020scaffold}.

\begin{figure}[!t] 
\centering\includegraphics[width=0.90\linewidth]{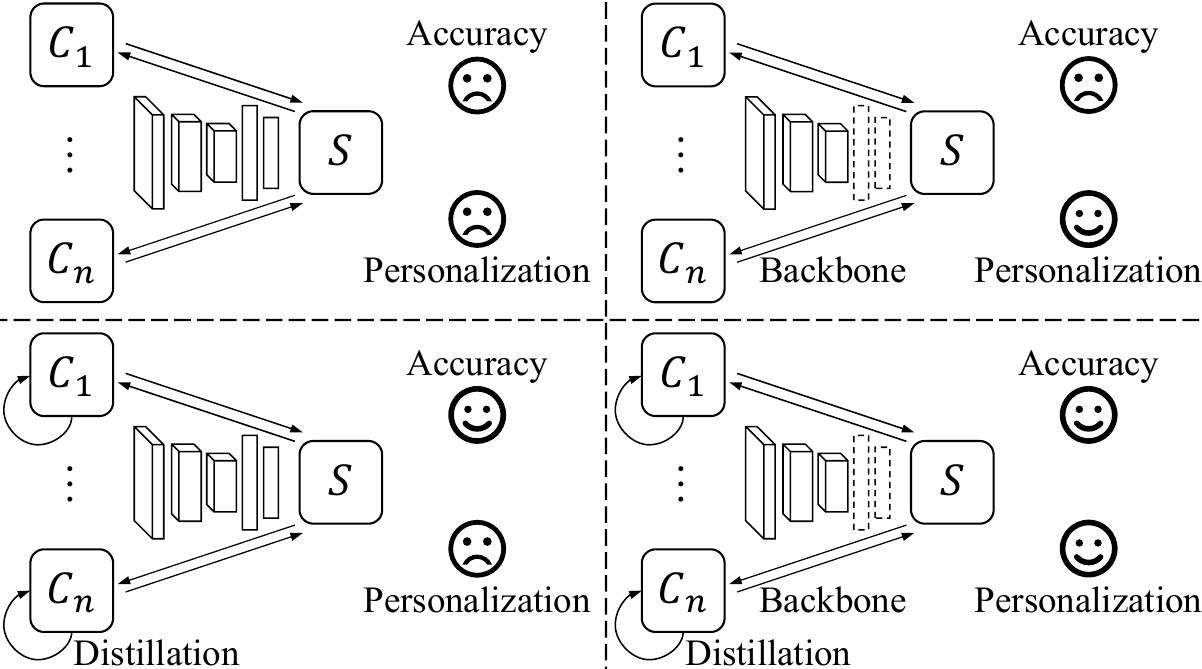}
\caption{To facilitate federated learning on heterogeneous data in practice, our proposed approach divides each client model into a shared backbone and a personalized head. Only the shared backbone is communicated between the client and the server. Furthermore, each client updates its model by employing self-distillation to mitigate accuracy degradation.}
\label{fig:approach_motivation} 
\end{figure}

Researchers have addressed this issue by focusing on model personalization~\cite{li2022learning}, catastrophic forgetting~\cite{lee2022preservation}, and related aspects. Recently, researchers have shown a preference for personalized federated learning~\cite{tan2022towards}, which can be implemented by adjusting model structures and model parameters. In terms of model structures, ~\cite{arivazhagan2019federated} proposes a personalization approach that divides the model into the same base layers and unique personalization layers. ~\cite{collins2021exploiting} aims to enable both clients and the server to learn a global representation while allowing each client to learn a unique head locally. However, these approaches lack optimal model performance when only uploading the partial network. In terms of model parameters, it primarily involves optimizing the local model based on global model parameters. The mainstream approaches include local fine-tuning~\cite{zhang2023fedala}, multi-task learning~\cite{cai2023many}, and knowledge distillation~\cite{jin2023personalized}. However, even though knowledge distillation facilitates knowledge transfer between the global and the local model, it neglects the adverse impact of the head network on personalization~\cite{oh2022fedbabu}, particularly when dealing with data heterogeneity. 

Upon reviewing the above research, as shown in Fig.~\ref{fig:approach_motivation}, traditional federated learning encounters substantial challenges within heterogeneous scenarios, adversely affecting both accuracy and personalization. To address these issues, personalization can be improved when only the shared backbone is communicated between the client and the server, but this improvement comes at the expense of accuracy. Conversely, the integration of knowledge distillation enhances accuracy while concurrently reducing personalization. 

In this work, we propose \textbf{FedBSD}, Personalized \textbf{Fed}erated Learning via \textbf{B}ackbone \textbf{S}elf-\textbf{D}istillation, which enhances the representational capability of the shared backbone while personalizing the model head. Our approach involves dividing each client model into a shared backbone and a private head. Communication between the client and the server is limited to the backbone weights, which are aggregated to establish a global backbone. By sharing partial knowledge in this manner, we achieve a common backbone and a personalized head. To enhance the common backbone, our approach utilizes self-distillation to transfer knowledge from the global backbone to the updated local model for each client. During each training round, the global backbone of the server serves as a teacher, and the local backbone of the client serves as a student. The client then performs distillation on its private data to update its model. This approach enhances client representations and ultimately improves overall performance.

In summary, this paper presents three main contributions. Firstly, our research reveals the constrained performance improvement of the common representation and emphasizes the necessity for optimization in federated learning. Secondly, we propose a personalized federated learning approach that integrates a personalized private head and enhances the representation of the shared backbone. Thirdly, we demonstrate the superiority of the proposed approach over existing mainstream personalized federated learning approaches through comprehensive experiments on simulated and real-world datasets.

\section{Related Work}

\myPara{Personalized federated learning.}~Federated learning is a collaborative training algorithm without exposing local private data~\cite{mcmahan2017communication}. However, it encounters challenges with data heterogeneity, leading to issues such as client drift. Consequently, researchers have focused on addressing the challenges posed by heterogeneous data in federated learning~\cite{haddadpour2021federated, karimireddy2020scaffold, zhang2020personalized, huang2022achieving, li2023effectiveness}. Personalized federated learning has emerged as a viable solution to address this issue and has been extensively investigated in recent research~\cite{tan2022towards}. Broadly, these approaches can be categorized into local fine-tuning~\cite{collins2022fedavg, setayesh2023perfedmask, zhang2023fedala}, meta-learning~\cite{khodak2019adaptive, fallah2020personalized, yang2023personalized}, multi-task learning~\cite{chen2022fedmsplit, ye2023pfedsa, zhang2023fedmpt}, and mixture of global and local models~\cite{deng2020adaptive, hanzely2020federated, mansour2020three}. These approaches personalize the parameters around the entire model. From the aspect of model architecture, ~\cite{liang2020think} shares only one part of the model while personalizing the rest of the model. ~\cite{arivazhagan2019federated} introduces FedPer to learn the base layers and the personalization layers. ~\cite{collins2021exploiting} proposes FedRep to learn the global representation and the unique head. As these approaches do not share all parameters, they only personalize one part of the model, while the other part is less personalized.

\myPara{Federated learning with knowledge distillation.}~Knowledge distillation is an algorithm proposed by~\cite{hinton2015distilling} that aims to enhance the performance of a small student model by leveraging a large teacher model. To extract essential information, researchers have started experimenting with integrating knowledge distillation into federated learning~\cite{mora2022knowledge}. This integration has the potential to reduce communication overhead and overcome challenges such as data heterogeneity. Federated distillation~\cite{jeong2018communication, li2019fedmd, wang2023dafkd} reduces the communication cost by transmitting the model logits with the data's label information instead of the model parameters. Alternatively, through distilling the local knowledge of the large model into the small model~\cite{wu2022communication}, only the parameters of the small model need to be communicated, thus reducing the communication overhead. Apart from the client-side knowledge distillation discussed earlier, some researchers also leverage multiple models on the server-side to carry out integrated knowledge distillation~\cite{lin2020ensemble, sui2020emnlp, cho2022heterogeneous}. Knowledge distillation can be a practical solution to address the issue of data heterogeneity in federated learning~\cite{lin2020ensemble, wen2022communication, wen2023communication}.
It can also be used to augment data for clients through data-free knowledge distillation~\cite{zhang2022fine, zhu2021data, chen2023best} and to facilitate knowledge transfer between global and local models~\cite{cho2022heterogeneous, usmanova2021distillation, ni2022esa}. Furthermore, it is an effective approach in federated learning for achieving personalization~\cite{ozkara2021quped, divi2021unifying, jin2023personalized}.

\section{Approach}

\begin{figure*}[!ht]
\centering
\includegraphics[width=0.80\linewidth]{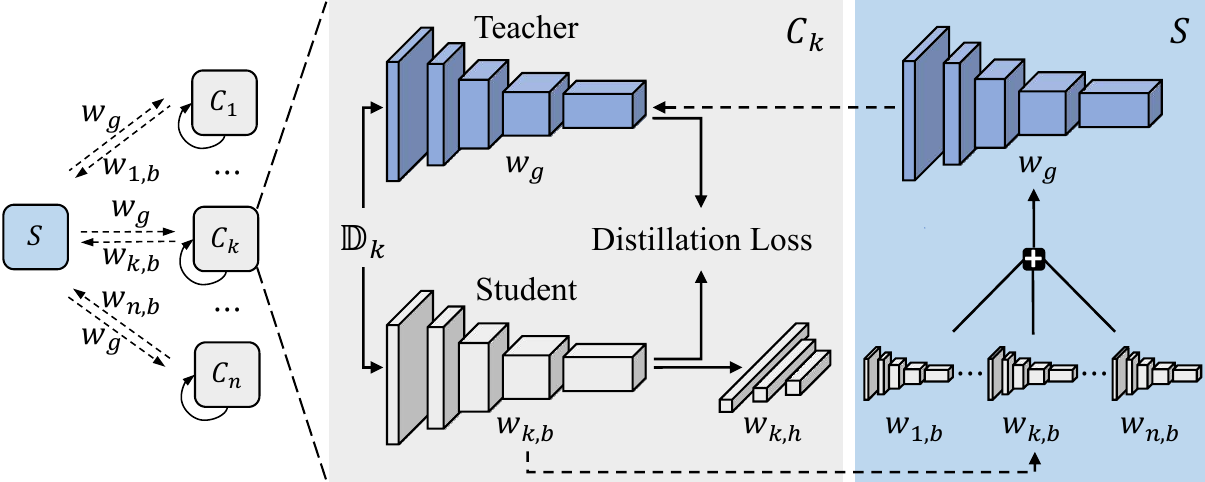} 
\caption{The framework of backbone self-distillation. Firstly, each client $C_k$ divides its local model $\bm{w}_k$ into a shared backbone $\bm{w}_{k,b}$ and a private head $\bm{w}_{k,h}$, and communicates only the shared backbone $\bm{w}_{k,b}$ with server $S$. Secondly, the server aggregates the shared backbone $\{\bm{w}_{k,b}\}_{k=1}^{n}$ to form a global backbone $\bm{w}_g$ and sends it back to each client. Finally, to mitigate accuracy degradation due to partial knowledge sharing, each client $C_k$ performs self-distillation on its local data $\mathbb{D}_k$ by transferring knowledge from the teacher $\bm{w}_{g}$ to the student $\bm{w}_{k,b}$, and trains the private head $\bm{w}_{k,h}$ to personalize the local model.}
\label{fig:approach_framework} 
\end{figure*}

Our backbone self-distillation approach aims to enhance the shared backbone and personalize the private head. As shown in Fig.~\ref{fig:approach_framework}, each client model is divided into a shared backbone and a private head, with only the backbone weights uploaded to the server for aggregation. This ensures that each client's head is unique and personalized. Purely personalizing the head for personalized federated learning has limitations. Therefore, we employ self-distillation to enhance the client model by using the global backbone as the teacher and the local backbone as the student. This procedure improves the performance of the local backbone.

\subsection{Problem Formulation}
Assume that there are $n$ clients, with local data $\mathbb{D}_{k}=\{\bm{x}_{k}, \bm{y}_{k}\}$, $k\in[1,\ldots,n]$, where $\bm{x}_{k}$ is the sample space, $\bm{y}_{k}$ is 
the label space, and $k$ denotes the $k$-th client. Each client usually has its local model $\phi_k(\bm{x}_{k};\bm{w}_{k})$ and the corresponding loss function $l_k$ , where $\bm{w}_{k}$ is the model weights of the $k$-th client. To utilize the data information from all clients, federated learning ensures privacy-preserving joint modeling by uploading the model weights or gradients. In the case of traditional federated learning, the format is as follows,
\begin{equation}
    \min\frac{1}{n}\sum\limits_{k=1}^{n}l_k(\phi_k(\bm{x}_{k};\bm{w}_{k}),\bm{y}_{k}).
\end{equation}

In order to minimize $l_k$, the $k$-th client uses local data $\mathbb{D}_{k}$ for training. Because of the limited availability of local data on the client, federated learning enables the continuous exchange of model information between the client and the server, thus enabling the client to utilize unseen data, thereby further reducing $l_k$.

Federated learning aims to aggregate the local models to obtain a powerful global model $\phi_g$. Although the global model performs well on average, its performance significantly deteriorates in heterogeneous environments. Some researchers contend that learning a global model may not be the optimal solution and propose the need for a personalized mechanism among heterogeneous clients.

\subsection{Common Representation Learning}

In traditional federated learning approaches, there is typically a requirement for exchanging all model parameters between the client and the server. Considering that each client is engaged in a classification task, we divide the client task into two parts: common representation learning and personalized classifier learning. Accordingly, for each client model $\phi_k(\bm{x}_{k};\bm{w}_{k})$, we divide it as follows, 
\begin{equation}
    \phi_k(\bm{x}_k;\bm{w}_k) \rightarrow \phi_k(\bm{x}_k;\bm{w}_{k,b},\bm{w}_{k,h}),
\end{equation}
where $\bm{w}_{k,b}$ and $\bm{w}_{k,h}$ correspond to the weights of the backbone and the head of the $k$-th client model, respectively. We only exchange $\bm{w}_{k,b}$ between the client and the server. Then, the server aggregates the shared backbone weights to obtain the global backbone,
\begin{equation}
    \label{aggregation}
    \bm{w}_{g} =\frac{1}{n}\sum\limits_{k=1}^{n} \bm{w}_{k,b}.
\end{equation}

After the server completes the aggregation of the shared backbone, the weights are returned to each client to replace the previous backbone, i.e., $\bm{w}_{k,b} = \bm{w}_{g}$. 

In our approach, the server aggregates the shared backbone to obtain a common representation. Meanwhile, each client utilizes local data to learn its personalized head. Upon reviewing the entire model, we observe that while the head is adequately personalized, the common representation is not sufficiently personalized. In the following section, we explore further optimization for the shared backbone to enhance the personalization of the representation.

\subsection{Backbone Self-distillation}

The common representation mitigates to some extent the harm caused by data heterogeneity in federated learning, but it is still essentially a general federated averaging approach. Clearly, for the $k$-th client with local data $\mathbb{D}_{k}$, local model $\phi_k(\bm{x}_{k};\bm{w}_{k})$ does not achieve the best performance. Therefore, we consider the optimization of the shared backbone on the client based on the previous section in order to attain improved personalized models.

Knowledge distillation is widely used in transfer learning to extract model knowledge effectively. Generally, it first trains a teacher model $\phi_t({\bm{x}};\bm{w}_t)$, and then allows the student model $\phi_s({\bm{x}};\bm{w}_s)$ to learn the probability distribution of output from the teacher model to improve the generalization.
\begin{equation}
    \label{KD_loss_model}
    \mathcal L =\ell(\phi_t(\bm{x};\bm{w}_t), \phi_s(\bm{x};\bm{w}_s)),
\end{equation}
where $\ell(\cdot)$ denotes the distillation loss.

Knowledge distillation includes two losses: the Cross-Entropy (CE) loss $\mathcal{L}_{\rm CE}$ and the Kullback–Leibler (KL) divergence loss $\mathcal{L}_{\rm KL}$.
\begin{equation}
\label{KD_loss_detail}
    \mathcal{L}=\mathcal{L}_{\rm CE}(\mathcal{O}_s, \bm{y}) + \lambda  \mathcal{L}_{\rm KL}(\mathcal{P}_{t}, \mathcal{P}_{s}),
\end{equation}
where $\mathcal{P}_{t}$ and $\mathcal{P}_{s}$ are the prediction of the teacher and student models, $\mathcal{O}_s$ is the output of the student model, $\bm{y}$ is the true label and $\lambda$ is the weight factor to balance the two losses.

For multi-class classification tasks with $m$ classes, The KL loss function is defined as follows,
\begin{equation}
\label{KL_loss}
    \mathcal{L}_{\rm KL}(\mathcal{P}_{t}, \mathcal{P}_{s})=-\sum\limits_{i=1}^{m}\mathcal{P}_{s,i}(\tau)\log(\frac{\mathcal{P}_{s,i}(\tau)}{\mathcal{P}_{t,i}(\tau)}),
\end{equation}
where $\mathcal{P}_{t,i}(\tau)$ and $\mathcal{P}_{s,i}(\tau)$ are the probabilistic prediction of the teacher and student models with $i$-th class at temperature $\tau$.
\begin{equation}
\label{softmax}
    \mathcal{P}_{t,i}(\tau)=\frac{\exp(\mathcal{O}_{t,i}/ \tau)}{\sum_{i=1}^m\exp(\mathcal{O}_{t,i} / \tau)},\mathcal{P}_{s,i}(\tau)=\frac{\exp(\mathcal{O}_{s,i}/ \tau)}{\sum_{i=1}^m\exp(\mathcal{O}_{s,i} / \tau)} ,
\end{equation}
where $\mathcal{O}_{t, i}$ and $\mathcal{O}_{s, i}$ are the output of the teacher and student models.

Currently, some researchers have implemented knowledge distillation in federated learning, particularly through self-distillation, where the teacher and student models are identical. However, most of these approaches require introducing additional data~\cite{zheng2023federated} and uploading all parameters from each client~\cite{jin2023personalized}, significantly increasing communication costs. 

In contrast, our backbone self-distillation does not require any additional data for distillation, and all clients only upload the shared backbone. This approach not only reduces communication costs but also guarantees the personalization of the private head.

For each client, we assign the global backbone $\phi_g(\bm{x}_{k}; \bm{w}_{g})$ as the teacher model and the local backbone $\phi_k(\bm{x}_{k}; \bm{w}_{k,b})$ as the student model. Therefore, total distillation loss is formulated as follows,
\begin{equation}
\label{KL_loss_FedBSD}
    \mathcal L =\ell(\phi_g(\bm{x}_{k}; \bm{w}_{g}), \phi_k(\bm{x}_{k}; \bm{w}_{k, b})).
\end{equation}

Finally, we compute the backbone self-distillation loss $\mathcal{L}_{\rm BSD}$, which replaces the traditional KL loss $\mathcal{L}_{\rm KL}$, and combine it with the CE loss $\mathcal{L}_{\rm CE}$ to formulate the total distillation loss as follows,
\begin{equation}
\label{FedBSD_loss}
    \mathcal{L}=\mathcal{L}_{\rm CE}(\mathcal{O}_s, \bm{y}) + \lambda  \mathcal{L}_{\rm BSD}(\mathcal{P}_{g}, \mathcal{P}_{b}),
\end{equation}
where $\mathcal{P}_{g}$ and $\mathcal{P}_{b}$ are the prediction of the global backbone and the local backbone.

As shown in Fig.~\ref{fig:approach_framework}, unlike traditional knowledge distillation, Because the model is divided into the backbone and the head, the objective of distillation here has changed to features rather than logits. The backbone is employed for uploading and implementing self-distillation to enhance personalization, whereas the head is exclusively trained using local data to guarantee personalization.

Alg.~\ref{algorithm:FedBSD} provides the training procedure. During each training round, a certain percentage of clients is randomly selected, the server distributes the global model to these clients, and each client divides the local model into a shared backbone and a private head. Next, we execute an SGD optimizer to train the head and perform self-distillation to train the backbone based on the local data, where in self-distillation training, the teacher model is the global backbone received from the server, and the student model is the original local backbone. Finally, each client uploads their local backbone weights, and the server aggregates them to obtain the global weights. 

\begin{algorithm}[!ht]
\caption{FedBSD}
\label{algorithm:FedBSD}
    \begin{algorithmic}[1]
        \Require Number of clients $n$, the $k$-th client local data $\mathbb{D}_{k}$, local head epochs $E_h$, local backbone epochs $E_b$, communication rounds $T$, learning rate $\eta$.
        \Ensure Personalized models $\bm{w}_{1}, \bm{w}_{2}, \ldots, \bm{w}_{n}$
        \State Initialization $\bm{w}^{0}_{g}$
        \For{each round $t=1,2,\ldots,T$}
            \State $K \leftarrow$ a random subset of the $n$ clients
            \State Server sends $\bm{w}^{t-1}_g$ to the selected clients

            \For{each client $k=1,2,\ldots,K$ }
                \State Client divides $\bm{w}^{t-1}_k$ into $(\bm{w}^{t-1}_{k,b}, \bm{w}^{t-1}_{k,h})$
                \For{local head epoch $e = 1,2,\ldots,E_h$} 
                    \State $\bm{w}^{t}_{k,h}=$ SGD$(\phi_{k}(\mathbb{D}_{k}; \bm{w}^{t-1}_{g}); \bm{w}^{t-1}_{k,h})$
                \EndFor
               
                \For{local backbone epoch $e = 1,2,\ldots,E_b$}
                    \State $\bm{w}^{t}_{k,b}=$ BSD$(\mathbb{D}_{k}; \bm{w}^{t-1}_{g}, \bm{w}^{t-1}_{k,b}, \bm{w}^{t}_{k,h})$ 
                \EndFor
                \State Client sends $\bm{w}^{t}_{k,b}$ to server
            \EndFor
            
            \State Server aggregates $\bm{w}^{t}_g=\frac{1}{K}\sum_{k=1}^{K}\bm{w}_{k,b}^{t}$ 
             
        \EndFor
        \State \Return $\bm{w}_{1}, \bm{w}_{2}, \ldots, \bm{w}_{n}$
        
        \Procedure{BSD}{$\mathbb{D}_{k}; \bm{w}_{g}, \bm{w}_{k,b}, \bm{w}_{k,h}$}
        \For{local data $\mathbb{D}_{k} = \{\bm{x}_{k}, \bm{y}_{k}\}$ }
            \State /* As detailed in Equation~\ref{FedBSD_loss} */
            \State $\bm{w}_{k,b} = \bm{w}_{k,b} - \eta \nabla \mathcal{L}(\bm{x}_{k}, \bm{y}_{k}; \bm{w}_g, \bm{w}_{k,b}, \bm{w}_{k,h})$
        \EndFor
        \State \Return $\bm{w}_{k,b}$
        \EndProcedure
    \end{algorithmic}
\end{algorithm}

\section{Experiments}
To validate the effectiveness of our approach, we conduct experiments on simulated and real-world datasets and compare with 12 state-of-the-art approaches.

\subsection{Experimental Setup}
\myPara{Datasets.}~We consider the image classification task and utilize three datasets for approach validation, i.e., CIFAR10~\cite{krizhevsky2009learning}, CIFAR100~\cite{krizhevsky2009learning}, and FEMNIST~\cite{caldas2018leaf}. To simulate a heterogeneous federated scenario, we randomly assign different numbers of classes $S$ to $N$ clients in a dataset. For CIFAR datasets, CIFAR10 has three settings $(N=100, S=2)$, $(N=100, S=5)$ and $(N=1000, S=2)$, CIFAR100 has two settings $(N=100, S=5)$ and $(N=100, S=20)$. For FEMNIST dataset, we limit the dataset only to include 10 handwritten letters and then distribute the samples to clients based on a log-normal distribution as in~\cite{collins2021exploiting}, and we consider the case of $(N=150, S=3)$, which means 150 clients with 3 classes per client.  

Furthermore, we have selected multiple datasets from transfer learning to evaluate the performance of the federated learning algorithm like~\cite{li2021fedbn}. Two real-world datasets, i.e., DomainNet~\cite{peng2019moment} and Digits, are used for this purpose. Each dataset serves as a client in the experimentation, making the clients heterogeneous. The DomainNet dataset consists of data from six distinct domains, including Infograph, Painting, Sketch, Clipart, Real, and Quickdraw. The Digits dataset consists of SVHN~\cite{netzer2011reading}, MNIST-M~\cite{ganin2015unsupervised}, SYNTH~\cite{ganin2015unsupervised},  MNIST~\cite{lecun1998gradient}, and USPS~\cite{hull1994database}.

\myPara{Models.}~We use two convolutional layers and three fully-connected layers for the CIFAR datasets, MLP with two hidden layers for the FEMNIST dataset, AlexNet~\cite{krizhevsky2012imagenet} for the DomainNet dataset, three convolutional layers and three fully-connected layers for the Digits dataset. To ensure fairness, we adhere to the setting described in \cite{collins2021exploiting} and designate the last layer of the model as the personalized head. For the other layers, we treat them as the shared backbone and apply self-distillation to enhance their personalized representations. 

\myPara{Baselines.}~We conduct comparisons with both non-personalized and personalized federated learning approaches. In the case of non-personalized federated learning, we compare with FedAvg~\cite{mcmahan2017communication}, FedProx~\cite{li2020federated}, SCAFFOLD~\cite{karimireddy2020scaffold}, and the results obtained after fine-tuning (FT) these approaches. FT results are obtained by conducting an additional 10 epochs of head training on each client, utilizing their local data, once the entire federated learning process is completed. In the case of personalized federated learning, we compare with PerFedAvg~\cite{fallah2020personalized}, Ditto~\cite{li2021ditto},  L2GD~\cite{hanzely2020federated}, APFL~\cite{deng2020adaptive}, LG-FedAvg~\cite{liang2020think}, FedPer~\cite{arivazhagan2019federated}, FedRep~\cite{collins2021exploiting}, PFedDS~\cite{jin2023personalized}, and FedPAC~\cite{xu2023personalized}. PerFedAvg is a meta-learning approach. Ditto is a multi-task learning approach. L2GD and APFL both seek an implicit mixture of global and local models. LG-FedAvg learns local representations and a global head. FedPer and FedRep both learn a global representation and personalized heads. PFedDS distills the knowledge of previous personalized models to current local models. FedPAC introduces feature alignment for enhancing representation and classifier combination for improving personalization.   

\myPara{Hyperparameters.}~In simulated experiments, only clients with rate $r=0.1$ are selected. The communication round for the CIFAR is $T=100$, while for the FEMNIST is $T=200$. Before communication, one round of backbone optimization is performed for the CIFAR10 case with $N=100$, and 5 rounds of backbone optimization is performed for all other cases. For FedRep and our FedBSD, the private head requires 10 rounds of updates. The results are calculated as the average of the last 10 communication rounds. In real-world experiments, the basic setup remains the same as described above. However, the local training epoch is $E=1$, and the participation rate of clients is $r=1$. Additionally, accuracy analysis is conducted for each client on the corresponding test dataset. 

The SGD optimizer with a momentum of $0.5$ is used, and the learning rate is set to $0.01$ for all approaches. In terms of the hyperparameters in knowledge distillation, the temperature is set to $\tau=2$, and the weight factor is set to $\lambda=1$, where $\lambda=1$ denotes that local training and global knowledge make an equal contribution.

\subsection{Experimental Results}

\myPara{Simulated experiments.}~We present the experimental results of FedBSD across three datasets and six experimental settings, as shown in Tab.~\ref{tab:result_simulated}. The experimental results demonstrate that our approach FedBSD not only outperforms existing federated learning approaches but also significantly surpasses the accuracy in the Local only case, where each client trains its model locally. In the CIFAR10 experiment with setup $(N=100, S=2)$, FedBSD achieves a performance that is less than 0.90\% away from the Local only results. These results affirm that FedBSD, with the utilization of backbone self-distillation, effectively addresses the issue of client drift in federated learning and achieves impressive performance. Notably, when using the MLP architecture on the FEMNIST dataset, FedBSD surpasses the performance of FedRep by 4.75\%. This finding suggests that FedBSD performs exceptionally well with simple MLP architecture when dealing with limited datasets, making it highly suitable for practical applications. Additionally, the poor performance of PFedSD and FedPAC further highlights the negative impact of the private head, thereby confirming the validity of our self-distillation approach for the backbone only.

\begin{table}[!ht]
    \caption{Average test accuracy of CIFAR10, CIFAR100, and FEMNIST for various settings. FT means fine-tuning.}
    \label{tab:result_simulated}
    \begin{center}
    \footnotesize
    \resizebox{\linewidth}{!}{
        \begin{tabular}{lcccccc}
        \toprule
        \multirow{2}{*}{Approach} & \multicolumn{3}{c}{{CIFAR10}} & \multicolumn{2}{c}{{CIFAR100}} & FEMNIST  \\
        \cmidrule(lr){2-4}\cmidrule(lr){5-6}\cmidrule(lr){7-7}
         & $(100,2)\!$ & $\!(100,5)$ & $(1000,2)$ & $(100,5)$ & $(100,20)$ & $(150,3)$ \\
        \midrule
        Local & $\mathbf{89.79}$ & 70.68 & 78.30 & 75.29 & 41.29 & 60.86 \\
        \midrule
        FedAvg \cite{mcmahan2017communication} &  42.65 & 51.78 & 44.31 & 23.94 & 31.97 & 51.64 \\
        FedAvg+FT & 87.65 & 73.68  & 82.04 & 79.34 & 55.44 & 72.41\\
        FedProx \cite{li2020federated} & 39.92 & 50.99 & 21.93  & 20.17 & 28.52 & 18.89 \\
        FedProx+FT & 85.81 & 72.75 & 75.41 & 78.52 & 55.09 & 53.54 \\
        SCAFFOLD \cite{karimireddy2020scaffold} & 37.72 & 47.33 & 33.79 & 20.32 & 22.52 & 17.65 \\
        SCAFFOLD+FT  & 86.35 & 68.23 & 78.24 & 78.88 & 44.34 & 52.11  \\
        \midrule 
        PerFedAvg \cite{fallah2020personalized} & 82.27 & 67.20 & 67.36 & 72.05 & 52.49 & 71.51 \\
        Ditto \cite{li2021ditto} & 85.39 & 70.34 & 80.36 & 78.91 & 56.34 & 68.28 \\
        L2GD \cite{hanzely2020federated} & 81.04 & 59.98 & 71.96 & 72.13 & 42.84 & 66.18  \\
        APFL \cite{deng2020adaptive} & 83.77 & 72.29 & 82.39 & 78.20 & 55.44 & 70.74 \\
        LG-FedAvg \cite{liang2020think} & 84.14  & 63.02 &  77.48 & 72.44 &38.76 & 62.08 \\
        FedPer \cite{arivazhagan2019federated} & 87.13 & 73.84 & 81.73 & 76.00 & 55.68 & 76.91 \\
        FedRep \cite{collins2021exploiting}& 87.70 & 75.68 & 83.27 & 79.15 & 56.10 & 78.56 \\
        PFedSD \cite{jin2023personalized}& 86.52 & 72.61 & 80.95 & 78.01 & 51.23 & 75.37 \\
        FedPAC \cite{xu2023personalized}& 87.80 & 73.22 & 82.74 & 79.10 & 53.32 & 77.93 \\
        \midrule
        FedBSD & $\mathbf{88.89}$ & $\mathbf{76.63}$ & $\mathbf{83.61}$ & $\mathbf{79.39}$ & $\mathbf{58.80}$ & $\mathbf{83.31}$ \\ 
        \bottomrule
        \end{tabular}
    }
\end{center}
\end{table}

\myPara{Real-world experiments.}~The use of local batch normalization (BN) before averaging the model helps alleviate feature bias in the heterogeneous federated scenario~\cite{li2021fedbn}. To ensure a fair comparison, we utilize AlexNet in the DomainNet dataset and six convolutional layers in the Digits dataset as the base models. Furthermore, a BN layer is added after each convolutional layer or fully-connected layer. Methodologically, we introduce FedBN~\cite{li2021fedbn}, which excludes the uploading of a BN layer in comparison to FedAvg. FedBN has demonstrated that excluding the BN layer contributes to the improvement of local models in federated learning. This claim is further supported by the experimental results. Consequently, we adopt this exclusion BN strategy in our approach FedBSD.

\begin{figure*}[htbp]
\centering 
    \includegraphics[width=0.49\linewidth]{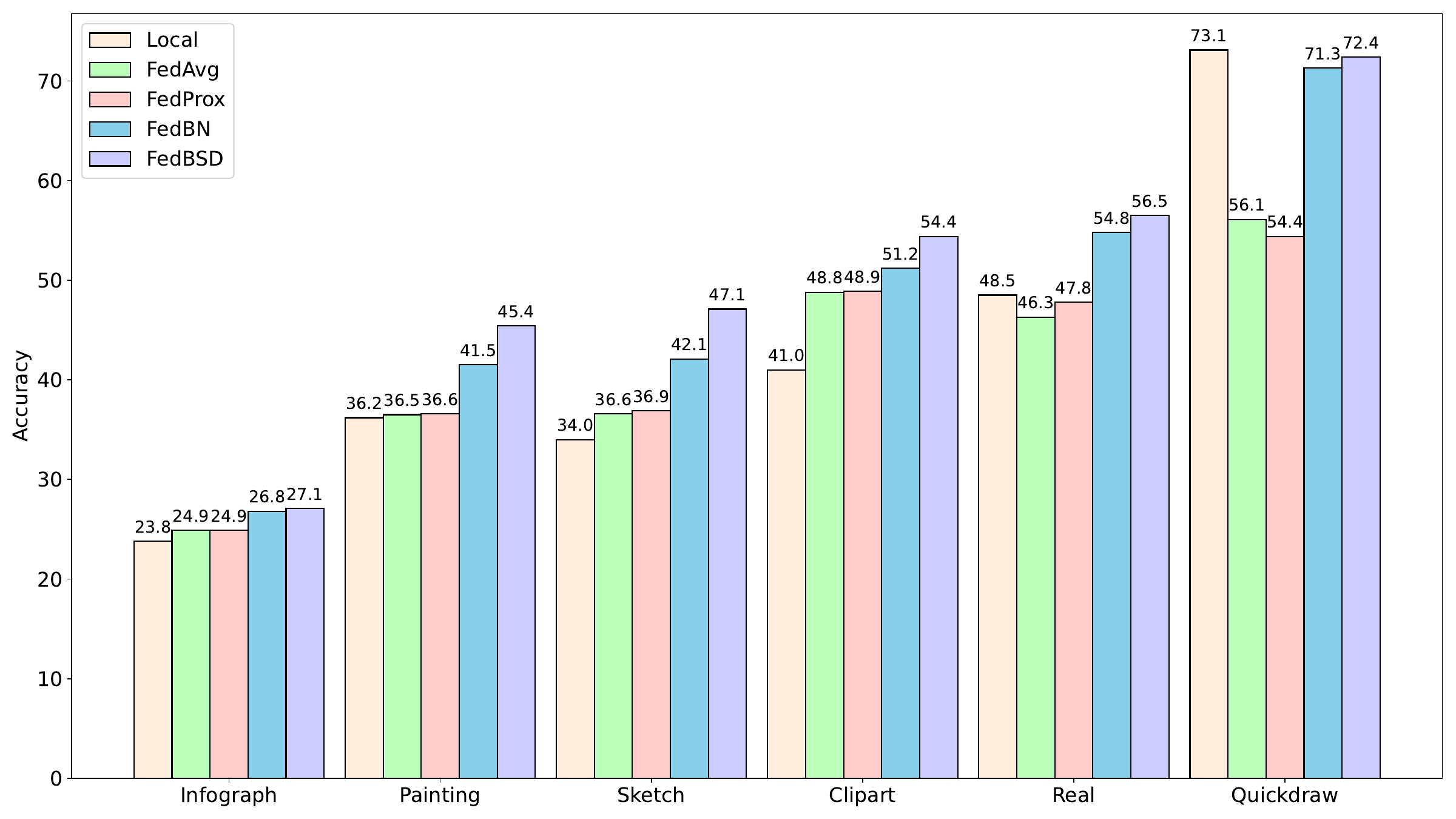}
    \includegraphics[width=0.49\linewidth]{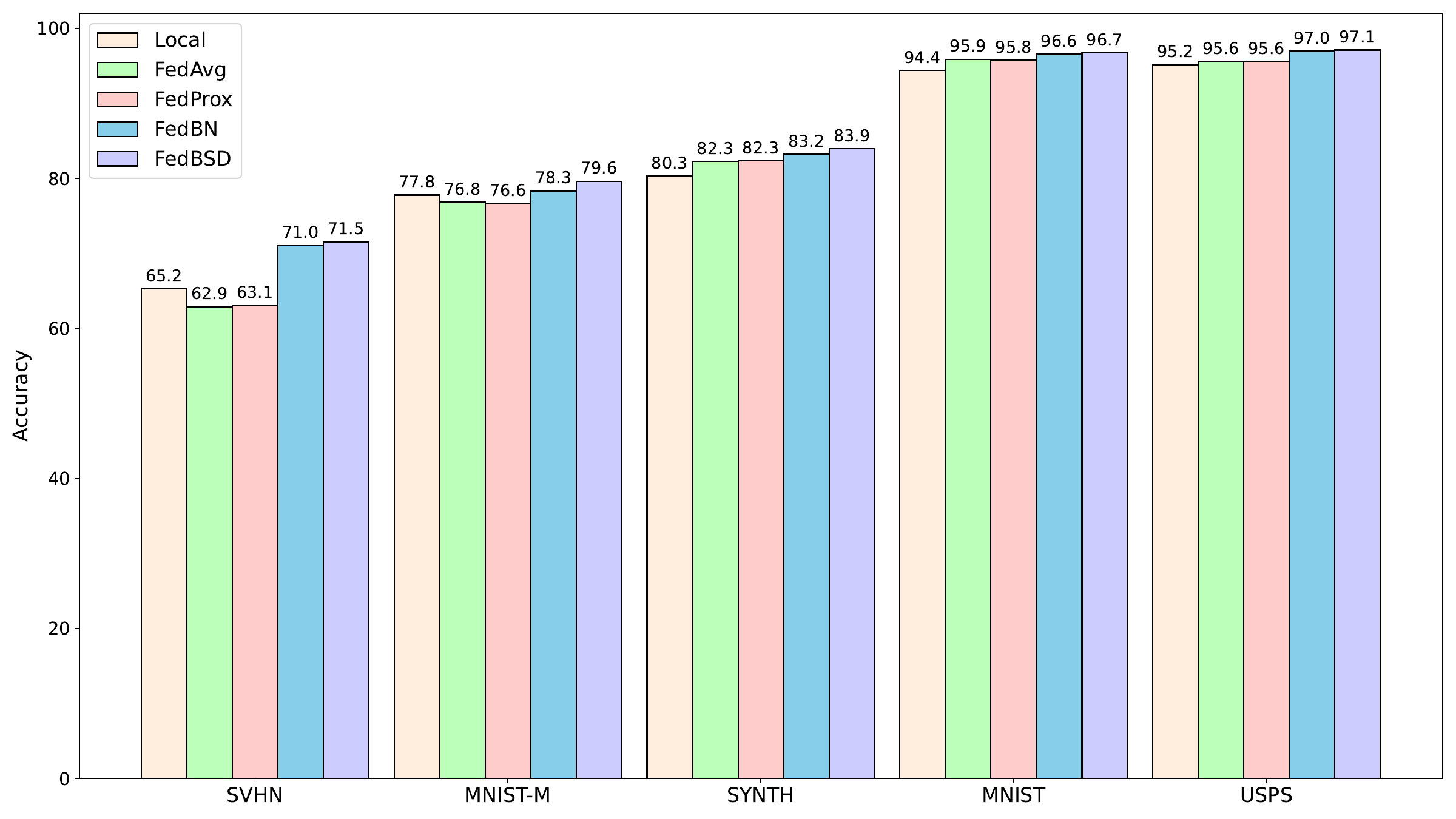}
    \caption{Test accuracy of various approaches on DomainNet (left) and Digits (right).} 
    \label{fig:real_world}
\end{figure*}

Fig.~\ref{fig:real_world} illustrates the accuracy achieved over 300 communication rounds, with each round consisting of one epoch on the client. It can be observed that our approach FedBSD demonstrates impressive performance on both the DomainNet and Digits datasets.

In the DomainNet experiment, it was observed that traditional approaches, FedAvg and FedProx, are ineffective in improving the performance of all clients. However, these approaches actually result in performance degradation due to interference from other clients. This implies that in a highly heterogeneous scenario, the presence of other clients becomes a source of noise that interferes with the local model of a specific client. In contrast, FedBSD demonstrates remarkable performance across all clients compared to FedBN.

In the Digits experiment, FedBSD also demonstrates a significant advantage over the clients, in contrast to other approaches. Furthermore, the communication parameters of FedBSD are substantially reduced as it eliminates the need to upload the BN layer and head layer. These findings strongly indicate that our approach successfully reduces communication parameters while simultaneously improving performance.

\begin{figure}[htbp]
\centering 
    \includegraphics[width=0.49\linewidth]{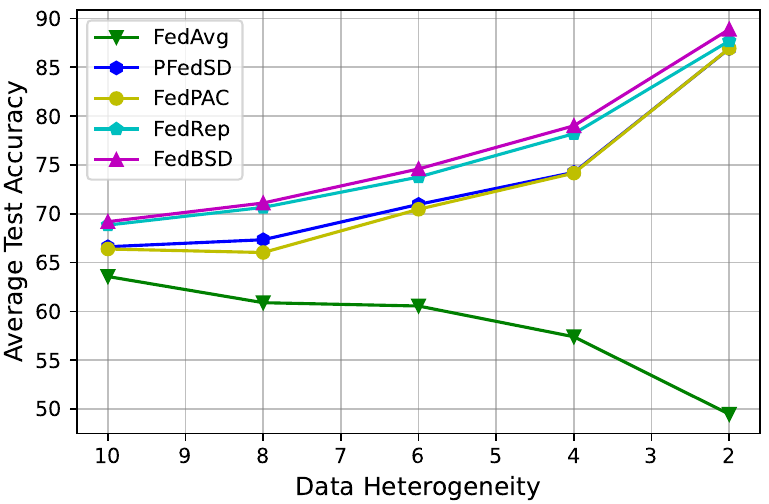}
    \includegraphics[width=0.49\linewidth]{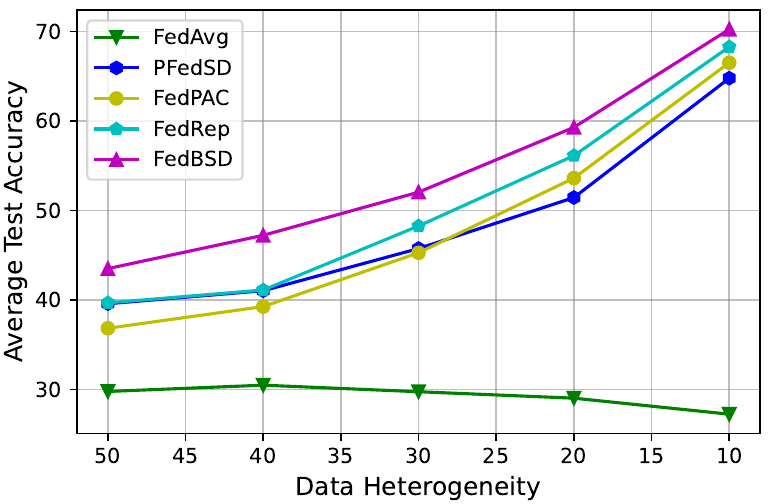}
    \includegraphics[width=0.49\linewidth]{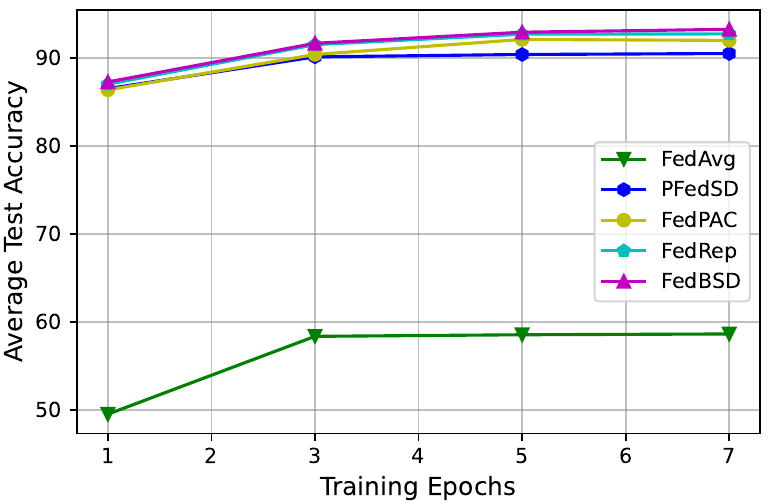}
    \includegraphics[width=0.49\linewidth]{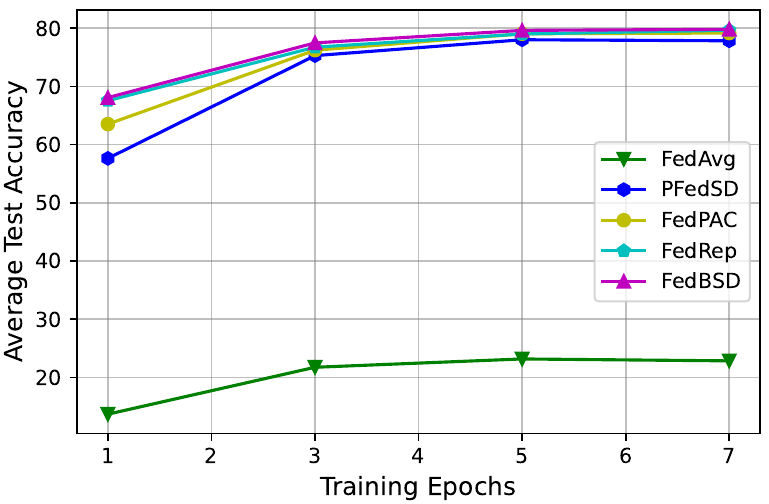}
    \includegraphics[width=0.49\linewidth]{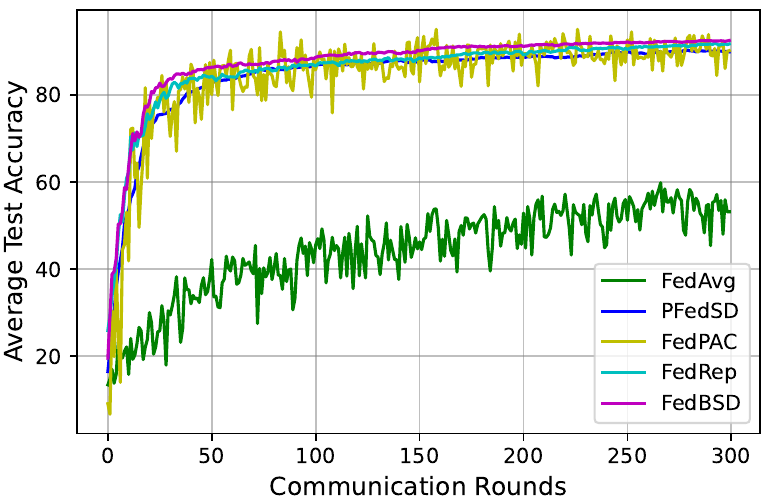}
    \includegraphics[width=0.49\linewidth]{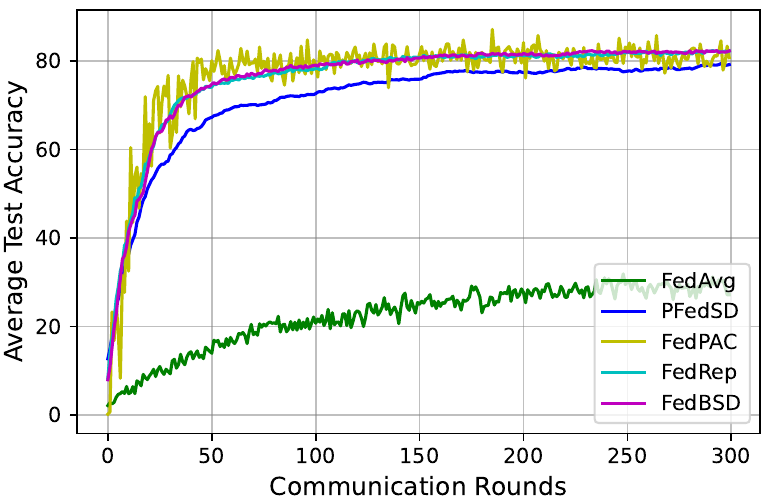}
    \caption{Ablation studies for data heterogeneity, training epochs, and communication rounds on CIFAR10 (left) and CIFAR100 (right).} 
    \label{fig:ablation_studies}
\end{figure}

\subsection{Ablation Studies}

This section aims to conduct ablation studies on the CIFAR10 and CIFAR100 datasets to investigate the effects of data heterogeneity, training epochs, and communication rounds. In the experimental setup, unless otherwise stated, the CIFAR10 dataset is set to $(N=100, S=2)$, and the CIFAR100 dataset is set to $(N=100, S=5)$.

\myPara{Effect of data heterogeneity.}~The total classes $S$ on each client serve as an indicator of the data heterogeneity, we compare the performance across different $S$ values. For the CIFAR10 dataset, the parameter $S$ is tuned among $\{2, 4, 6, 8, 10\}$. For the CIFAR100 dataset, the parameter $S$ is tuned among $\{10, 20, 30, 40, 50\}$. When $S=10$, CIFAR10 is evenly distributed, and each client contains data from 10 classes. This homogeneity ensures that each client has sufficient information to learn from the other  clients. As $S$ decreases, the data becomes more heterogeneous. For instance, when $S=2$, each client contains only two classes. Fig.~\ref{fig:ablation_studies} shows that FedBSD outperforms other approaches in all heterogeneous conditions. As data heterogeneity increases, the performance of FedAvg progressively declines, while the opposite trend is observed for FedBSD. This occurs because the presence of other clients can interfere with the local model, acting as noise and resulting in degraded model performance. Most of the real-world scenarios exhibit heterogeneity, with a small value for $S$, making our approach FedBSD highly suitable.

\myPara{Effect of training epochs.}~The number of local training epochs $E$ on the client is also an essential indicator, which refers to the backbone training epochs, specifically in FedRep and FedBSD. For all datasets, the parameter $E$ is tuned among $\{1, 3, 5, 7\}$. If higher accuracy can be achieved with fewer training epochs, there is no necessity to allocate excessive training time and computational resources to train the client, thereby improving the training efficiency of federated learning. Fig.~\ref{fig:ablation_studies} demonstrates that when $E=5$, the model accuracy is already significantly high. Moreover, increasing the number of training epochs has minimal impact on improving accuracy. Instead, it hampers training efficiency by increasing the training time and computational costs.

\myPara{Effect of communication rounds.}~ Communication efficiency in federated learning depends on the number of communication parameters and rounds. While we have previously analyzed FedBSD's low communication parameters, we now turn our attention to analyzing the number of communication rounds $T$. Fig.~\ref{fig:ablation_studies} shows the personalized accuracy in each round during training. FedBSD shows faster and smoother convergence and consistently outperforms other approaches. Therefore, FedBSD achieves the highest accuracy with the fewest number of communication rounds. Furthermore, we observe that the accuracy of FedAvg and FedPAC exhibit instability, indicating the challenges in achieving convergence when dealing with data heterogeneity.

% All these results provide additional evidence that our approach FedBSD exhibits outstanding performance in all aspects.

\section{Conclusion}
In this paper, we propose a backbone self-distillation approach called FedBSD for personalized federated learning. FedBSD divides each client model into a shared backbone and a private head, where the backbone is communicated with the server for knowledge sharing among clients. Each client conducts self-distillation to transfer the global backbone knowledge, resulting in a personalized private head when enhancing accuracy. Importantly, the proposed approach does not rely on external data and can be readily extended to other federated learning approaches. In future work, we aim to combine backbone self-distillation with model compression to minimize communication costs.

\begin{acks} 
This work is partially supported by grants from the National Key Research and Development Program of China (2020AAA0140001), and the Beijing Natural Science Foundation (19L2040).
\end{acks}

%\newpage
\bibliographystyle{ACM-Reference-Format}
\bibliography{main}

%%% -*-BibTeX-*-
%%% Do NOT edit. File created by BibTeX with style
%%% ACM-Reference-Format-Journals [18-Jan-2012].

\begin{thebibliography}{65}

%%% ====================================================================
%%% NOTE TO THE USER: you can override these defaults by providing
%%% customized versions of any of these macros before the \bibliography
%%% command.  Each of them MUST provide its own final punctuation,
%%% except for \shownote{}, \showDOI{}, and \showURL{}.  The latter two
%%% do not use final punctuation, in order to avoid confusing it with
%%% the Web address.
%%%
%%% To suppress output of a particular field, define its macro to expand
%%% to an empty string, or better, \unskip, like this:
%%%
%%% \newcommand{\showDOI}[1]{\unskip}   % LaTeX syntax
%%%
%%% \def \showDOI #1{\unskip}           % plain TeX syntax
%%%
%%% ====================================================================

\ifx \showCODEN    \undefined \def \showCODEN     #1{\unskip}     \fi
\ifx \showDOI      \undefined \def \showDOI       #1{#1}\fi
\ifx \showISBNx    \undefined \def \showISBNx     #1{\unskip}     \fi
\ifx \showISBNxiii \undefined \def \showISBNxiii  #1{\unskip}     \fi
\ifx \showISSN     \undefined \def \showISSN      #1{\unskip}     \fi
\ifx \showLCCN     \undefined \def \showLCCN      #1{\unskip}     \fi
\ifx \shownote     \undefined \def \shownote      #1{#1}          \fi
\ifx \showarticletitle \undefined \def \showarticletitle #1{#1}   \fi
\ifx \showURL      \undefined \def \showURL       {\relax}        \fi
% The following commands are used for tagged output and should be
% invisible to TeX
\providecommand\bibfield[2]{#2}
\providecommand\bibinfo[2]{#2}
\providecommand\natexlab[1]{#1}
\providecommand\showeprint[2][]{arXiv:#2}

\bibitem[Arivazhagan et~al\mbox{.}(2019)]%
        {arivazhagan2019federated}
\bibfield{author}{\bibinfo{person}{Manoj~Ghuhan Arivazhagan}, \bibinfo{person}{Vinay Aggarwal}, \bibinfo{person}{Aaditya~Kumar Singh}, {et~al\mbox{.}}} \bibinfo{year}{2019}\natexlab{}.
\newblock \showarticletitle{Federated Learning with Personalization Layers}.
\newblock \bibinfo{journal}{\emph{arXiv:1912.00818}} (\bibinfo{year}{2019}).
\newblock


\bibitem[Cai et~al\mbox{.}(2023)]%
        {cai2023many}
\bibfield{author}{\bibinfo{person}{Ruisi Cai}, \bibinfo{person}{Xiaohan Chen}, \bibinfo{person}{Shiwei Liu}, {et~al\mbox{.}}} \bibinfo{year}{2023}\natexlab{}.
\newblock \showarticletitle{Many-Task Federated Learning: A New Problem Setting and a Simple Baseline}. In \bibinfo{booktitle}{\emph{CVPR}}. \bibinfo{pages}{5036--5044}.
\newblock


\bibitem[Caldas et~al\mbox{.}(2018)]%
        {caldas2018leaf}
\bibfield{author}{\bibinfo{person}{Sebastian Caldas}, \bibinfo{person}{Sai Meher~Karthik Duddu}, \bibinfo{person}{Peter Wu}, {et~al\mbox{.}}} \bibinfo{year}{2018}\natexlab{}.
\newblock \showarticletitle{LEAF: A Benchmark for Federated Settings}.
\newblock \bibinfo{journal}{\emph{arXiv:1812.01097}} (\bibinfo{year}{2018}).
\newblock


\bibitem[Chatterjee et~al\mbox{.}(2023)]%
        {chatterjee2023use}
\bibfield{author}{\bibinfo{person}{Pushpita Chatterjee}, \bibinfo{person}{Debashis Das}, {and} \bibinfo{person}{Danda~B Rawat}.} \bibinfo{year}{2023}\natexlab{}.
\newblock \showarticletitle{Use of Federated Learning and Blockchain towards Securing Financial Services}.
\newblock \bibinfo{journal}{\emph{arXiv:2303.12944}} (\bibinfo{year}{2023}).
\newblock


\bibitem[Chen et~al\mbox{.}(2023)]%
        {chen2023best}
\bibfield{author}{\bibinfo{person}{Huancheng Chen}, \bibinfo{person}{Chaining Wang}, {and} \bibinfo{person}{Haris Vikalo}.} \bibinfo{year}{2023}\natexlab{}.
\newblock \showarticletitle{The Best of Both Worlds: Accurate Global and Personalized Models through Federated Learning with Data-Free Hyper-Knowledge Distillation}. In \bibinfo{booktitle}{\emph{ICLR}}.
\newblock


\bibitem[Chen and Zhang(2022)]%
        {chen2022fedmsplit}
\bibfield{author}{\bibinfo{person}{Jiayi Chen} {and} \bibinfo{person}{Aidong Zhang}.} \bibinfo{year}{2022}\natexlab{}.
\newblock \showarticletitle{FedMSplit: Correlation-Adaptive Federated Multi-Task Learning across Multimodal Split Networks}. In \bibinfo{booktitle}{\emph{SIGKDD}}. \bibinfo{pages}{87--96}.
\newblock


\bibitem[Cho et~al\mbox{.}(2022)]%
        {cho2022heterogeneous}
\bibfield{author}{\bibinfo{person}{Yae~Jee Cho}, \bibinfo{person}{Andre Manoel}, \bibinfo{person}{Gauri Joshi}, {et~al\mbox{.}}} \bibinfo{year}{2022}\natexlab{}.
\newblock \showarticletitle{Heterogeneous Ensemble Knowledge Transfer for Training Large Models in Federated Learning}. In \bibinfo{booktitle}{\emph{IJCAI}}. \bibinfo{pages}{2881--2887}.
\newblock


\bibitem[Collins et~al\mbox{.}(2021)]%
        {collins2021exploiting}
\bibfield{author}{\bibinfo{person}{Liam Collins}, \bibinfo{person}{Hamed Hassani}, \bibinfo{person}{Aryan Mokhtari}, {et~al\mbox{.}}} \bibinfo{year}{2021}\natexlab{}.
\newblock \showarticletitle{Exploiting Shared Representations for Personalized Federated Learning}. In \bibinfo{booktitle}{\emph{ICML}}. \bibinfo{pages}{2089--2099}.
\newblock


\bibitem[Collins et~al\mbox{.}(2022)]%
        {collins2022fedavg}
\bibfield{author}{\bibinfo{person}{Liam Collins}, \bibinfo{person}{Hamed Hassani}, \bibinfo{person}{Aryan Mokhtari}, {et~al\mbox{.}}} \bibinfo{year}{2022}\natexlab{}.
\newblock \showarticletitle{FedAvg with Fine Tuning: Local Updates Lead to Representation Learning}. In \bibinfo{booktitle}{\emph{NeurIPS}}. \bibinfo{pages}{10572--10586}.
\newblock


\bibitem[Deng et~al\mbox{.}(2020)]%
        {deng2020adaptive}
\bibfield{author}{\bibinfo{person}{Yuyang Deng}, \bibinfo{person}{Mohammad~Mahdi Kamani}, {and} \bibinfo{person}{Mehrdad Mahdavi}.} \bibinfo{year}{2020}\natexlab{}.
\newblock \showarticletitle{Adaptive Personalized Federated Learning}.
\newblock \bibinfo{journal}{\emph{arXiv:2003.13461}} (\bibinfo{year}{2020}).
\newblock


\bibitem[Divi et~al\mbox{.}(2021)]%
        {divi2021unifying}
\bibfield{author}{\bibinfo{person}{Siddharth Divi}, \bibinfo{person}{Habiba Farrukh}, {and} \bibinfo{person}{Berkay Celik}.} \bibinfo{year}{2021}\natexlab{}.
\newblock \showarticletitle{Unifying Distillation with Personalization in Federated Learning}.
\newblock \bibinfo{journal}{\emph{arXiv:2105.15191}} (\bibinfo{year}{2021}).
\newblock


\bibitem[Fallah et~al\mbox{.}(2020)]%
        {fallah2020personalized}
\bibfield{author}{\bibinfo{person}{Alireza Fallah}, \bibinfo{person}{Aryan Mokhtari}, {and} \bibinfo{person}{Asuman Ozdaglar}.} \bibinfo{year}{2020}\natexlab{}.
\newblock \showarticletitle{Personalized Federated Learning with Theoretical Guarantees: {A} Model-Agnostic Meta-Learning Approach}. In \bibinfo{booktitle}{\emph{NeurIPS}}. \bibinfo{pages}{3557--3568}.
\newblock


\bibitem[Ganin and Lempitsky(2015)]%
        {ganin2015unsupervised}
\bibfield{author}{\bibinfo{person}{Yaroslav Ganin} {and} \bibinfo{person}{Victor Lempitsky}.} \bibinfo{year}{2015}\natexlab{}.
\newblock \showarticletitle{Unsupervised Domain Adaptation by Backpropagation}. In \bibinfo{booktitle}{\emph{ICML}}. \bibinfo{pages}{1180--1189}.
\newblock


\bibitem[Haddadpour et~al\mbox{.}(2021)]%
        {haddadpour2021federated}
\bibfield{author}{\bibinfo{person}{Farzin Haddadpour}, \bibinfo{person}{Mohammad~Mahdi Kamani}, \bibinfo{person}{Aryan Mokhtari}, {et~al\mbox{.}}} \bibinfo{year}{2021}\natexlab{}.
\newblock \showarticletitle{Federated Learning with Compression: Unified Analysis and Sharp Guarantees}. In \bibinfo{booktitle}{\emph{AISTATS}}. \bibinfo{pages}{2350--2358}.
\newblock


\bibitem[Hanzely and Richt{\'a}rik(2020)]%
        {hanzely2020federated}
\bibfield{author}{\bibinfo{person}{Filip Hanzely} {and} \bibinfo{person}{Peter Richt{\'a}rik}.} \bibinfo{year}{2020}\natexlab{}.
\newblock \showarticletitle{Federated Learning of a Mixture of Global and Local Models}.
\newblock \bibinfo{journal}{\emph{arXiv:2002.05516}} (\bibinfo{year}{2020}).
\newblock


\bibitem[Hinton et~al\mbox{.}(2015)]%
        {hinton2015distilling}
\bibfield{author}{\bibinfo{person}{Geoffrey Hinton}, \bibinfo{person}{Oriol Vinyals}, {and} \bibinfo{person}{Jeff Dean}.} \bibinfo{year}{2015}\natexlab{}.
\newblock \showarticletitle{Distilling the Knowledge in a Neural Network}.
\newblock \bibinfo{journal}{\emph{arXiv:1503.02531}} (\bibinfo{year}{2015}).
\newblock


\bibitem[Huang et~al\mbox{.}(2022)]%
        {huang2022achieving}
\bibfield{author}{\bibinfo{person}{Tiansheng Huang}, \bibinfo{person}{Shiwei Liu}, \bibinfo{person}{Li Shen}, {et~al\mbox{.}}} \bibinfo{year}{2022}\natexlab{}.
\newblock \showarticletitle{Achieving Personalized Federated Learning with Sparse Local Models}.
\newblock \bibinfo{journal}{\emph{arXiv:2201.11380}} (\bibinfo{year}{2022}).
\newblock


\bibitem[Hull(1994)]%
        {hull1994database}
\bibfield{author}{\bibinfo{person}{Jonathan~J. Hull}.} \bibinfo{year}{1994}\natexlab{}.
\newblock \showarticletitle{A Database for Handwritten Text Recognition Research}.
\newblock \bibinfo{journal}{\emph{IEEE TPAMI}} \bibinfo{volume}{16}, \bibinfo{number}{5} (\bibinfo{year}{1994}), \bibinfo{pages}{550--554}.
\newblock


\bibitem[Jeong et~al\mbox{.}(2018)]%
        {jeong2018communication}
\bibfield{author}{\bibinfo{person}{Eunjeong Jeong}, \bibinfo{person}{Seungeun Oh}, \bibinfo{person}{Hyesung Kim}, {et~al\mbox{.}}} \bibinfo{year}{2018}\natexlab{}.
\newblock \showarticletitle{Communication-Efficient On-Device Machine Learning: Federated Distillation and Augmentation under Non-IID Private Data}.
\newblock \bibinfo{journal}{\emph{arXiv:1811.11479}} (\bibinfo{year}{2018}).
\newblock


\bibitem[Jin et~al\mbox{.}(2023)]%
        {jin2023personalized}
\bibfield{author}{\bibinfo{person}{Hai Jin}, \bibinfo{person}{Dongshan Bai}, \bibinfo{person}{Dezhong Yao}, {et~al\mbox{.}}} \bibinfo{year}{2023}\natexlab{}.
\newblock \showarticletitle{Personalized Edge Intelligence via Federated Self-Knowledge Distillation}.
\newblock \bibinfo{journal}{\emph{IEEE TPDS}} \bibinfo{volume}{34}, \bibinfo{number}{2} (\bibinfo{year}{2023}), \bibinfo{pages}{567--580}.
\newblock


\bibitem[Kairouz et~al\mbox{.}(2021)]%
        {kairouz2019advances}
\bibfield{author}{\bibinfo{person}{Peter Kairouz}, \bibinfo{person}{H~Brendan McMahan}, \bibinfo{person}{Brendan Avent}, {et~al\mbox{.}}} \bibinfo{year}{2021}\natexlab{}.
\newblock \showarticletitle{Advances and Open Problems in Federated Learning}.
\newblock \bibinfo{journal}{\emph{Foundations and Trends in Machine Learning}} \bibinfo{volume}{14}, \bibinfo{number}{1-2} (\bibinfo{year}{2021}), \bibinfo{pages}{1--210}.
\newblock


\bibitem[Karargyris et~al\mbox{.}(2023)]%
        {karargyris2023federated}
\bibfield{author}{\bibinfo{person}{Alexandros Karargyris}, \bibinfo{person}{Renato Umeton}, \bibinfo{person}{Micah~J Sheller}, {et~al\mbox{.}}} \bibinfo{year}{2023}\natexlab{}.
\newblock \showarticletitle{Federated Benchmarking of Medical Artificial Intelligence with MedPerf}.
\newblock \bibinfo{journal}{\emph{Nature Machine Intelligence}}  \bibinfo{volume}{5} (\bibinfo{year}{2023}), \bibinfo{pages}{799–810}.
\newblock


\bibitem[Karimireddy et~al\mbox{.}(2020)]%
        {karimireddy2020scaffold}
\bibfield{author}{\bibinfo{person}{Sai~Praneeth Karimireddy}, \bibinfo{person}{Satyen Kale}, \bibinfo{person}{Mehryar Mohri}, {et~al\mbox{.}}} \bibinfo{year}{2020}\natexlab{}.
\newblock \showarticletitle{SCAFFOLD: Stochastic Controlled Averaging for Federated Learning}. In \bibinfo{booktitle}{\emph{ICML}}. \bibinfo{pages}{5132--5143}.
\newblock


\bibitem[Khodak et~al\mbox{.}(2019)]%
        {khodak2019adaptive}
\bibfield{author}{\bibinfo{person}{Mikhail Khodak}, \bibinfo{person}{Maria-Florina~F Balcan}, {and} \bibinfo{person}{Ameet~S Talwalkar}.} \bibinfo{year}{2019}\natexlab{}.
\newblock \showarticletitle{Adaptive Gradient-Based Meta-Learning Methods}. In \bibinfo{booktitle}{\emph{NeurIPS}}. \bibinfo{pages}{5917–5928}.
\newblock


\bibitem[Krizhevsky and Hinton(2009)]%
        {krizhevsky2009learning}
\bibfield{author}{\bibinfo{person}{Alex Krizhevsky} {and} \bibinfo{person}{Geoffrey Hinton}.} \bibinfo{year}{2009}\natexlab{}.
\newblock \showarticletitle{Learning Multiple Layers of Features from Tiny Images}.
\newblock  (\bibinfo{year}{2009}).
\newblock


\bibitem[Krizhevsky et~al\mbox{.}(2012)]%
        {krizhevsky2012imagenet}
\bibfield{author}{\bibinfo{person}{Alex Krizhevsky}, \bibinfo{person}{Ilya Sutskever}, {and} \bibinfo{person}{Geoffrey~E Hinton}.} \bibinfo{year}{2012}\natexlab{}.
\newblock \showarticletitle{ImageNet Classification with Deep Convolutional Neural Networks}. In \bibinfo{booktitle}{\emph{NeurIPS}}. \bibinfo{pages}{1106--1114}.
\newblock


\bibitem[LeCun et~al\mbox{.}(1998)]%
        {lecun1998gradient}
\bibfield{author}{\bibinfo{person}{Yann LeCun}, \bibinfo{person}{L{\'e}on Bottou}, \bibinfo{person}{Yoshua Bengio}, {et~al\mbox{.}}} \bibinfo{year}{1998}\natexlab{}.
\newblock \showarticletitle{Gradient-Based Learning Applied to Document Recognition}.
\newblock \bibinfo{journal}{\emph{Proc. IEEE}} \bibinfo{volume}{86}, \bibinfo{number}{11} (\bibinfo{year}{1998}), \bibinfo{pages}{2278--2324}.
\newblock


\bibitem[Lee et~al\mbox{.}(2022)]%
        {lee2022preservation}
\bibfield{author}{\bibinfo{person}{Gihun Lee}, \bibinfo{person}{Minchan Jeong}, \bibinfo{person}{Yongjin Shin}, {et~al\mbox{.}}} \bibinfo{year}{2022}\natexlab{}.
\newblock \showarticletitle{Preservation of the Global Knowledge by Not-True Distillation in Federated Learning}. In \bibinfo{booktitle}{\emph{NeurIPS}}. \bibinfo{pages}{38461--38474}.
\newblock


\bibitem[Li et~al\mbox{.}(2023)]%
        {li2023effectiveness}
\bibfield{author}{\bibinfo{person}{Bo Li}, \bibinfo{person}{Mikkel~N Schmidt}, \bibinfo{person}{Tommy~S Alstr{\o}m}, {et~al\mbox{.}}} \bibinfo{year}{2023}\natexlab{}.
\newblock \showarticletitle{On the Effectiveness of Partial Variance Reduction in Federated Learning With Heterogeneous Data}. In \bibinfo{booktitle}{\emph{CVPR}}. \bibinfo{pages}{3964--3973}.
\newblock


\bibitem[Li and Wang(2019)]%
        {li2019fedmd}
\bibfield{author}{\bibinfo{person}{Daliang Li} {and} \bibinfo{person}{Junpu Wang}.} \bibinfo{year}{2019}\natexlab{}.
\newblock \showarticletitle{FedMD: Heterogenous Federated Learning via Model Distillation}.
\newblock \bibinfo{journal}{\emph{arXiv:1910.03581}} (\bibinfo{year}{2019}).
\newblock


\bibitem[Li et~al\mbox{.}(2020b)]%
        {li2020invisiblefl}
\bibfield{author}{\bibinfo{person}{Qiushi Li}, \bibinfo{person}{Wenwu Zhu}, \bibinfo{person}{Chao Wu}, {et~al\mbox{.}}} \bibinfo{year}{2020}\natexlab{b}.
\newblock \showarticletitle{InvisibleFL: Federated Learning over Non-Informative Intermediate Updates against Multimedia Privacy Leakages}. In \bibinfo{booktitle}{\emph{ACM MM}}. \bibinfo{pages}{753--762}.
\newblock


\bibitem[Li et~al\mbox{.}(2022)]%
        {li2022learning}
\bibfield{author}{\bibinfo{person}{Shuangtong Li}, \bibinfo{person}{Tianyi Zhou}, \bibinfo{person}{Xinmei Tian}, {et~al\mbox{.}}} \bibinfo{year}{2022}\natexlab{}.
\newblock \showarticletitle{Learning to Collaborate in Decentralized Learning of Personalized Models}. In \bibinfo{booktitle}{\emph{CVPR}}. \bibinfo{pages}{9766--9775}.
\newblock


\bibitem[Li et~al\mbox{.}(2021a)]%
        {li2021ditto}
\bibfield{author}{\bibinfo{person}{Tian Li}, \bibinfo{person}{Shengyuan Hu}, \bibinfo{person}{Ahmad Beirami}, {et~al\mbox{.}}} \bibinfo{year}{2021}\natexlab{a}.
\newblock \showarticletitle{Ditto: Fair and Robust Federated Learning through Personalization}. In \bibinfo{booktitle}{\emph{ICML}}. \bibinfo{pages}{6357--6368}.
\newblock


\bibitem[Li et~al\mbox{.}(2020a)]%
        {li2020federated}
\bibfield{author}{\bibinfo{person}{Tian Li}, \bibinfo{person}{Anit~Kumar Sahu}, \bibinfo{person}{Manzil Zaheer}, {et~al\mbox{.}}} \bibinfo{year}{2020}\natexlab{a}.
\newblock \showarticletitle{Federated Optimization in Heterogeneous Networks}. In \bibinfo{booktitle}{\emph{MLSys}}. \bibinfo{pages}{429--450}.
\newblock


\bibitem[Li et~al\mbox{.}(2021b)]%
        {li2021fedbn}
\bibfield{author}{\bibinfo{person}{Xiaoxiao Li}, \bibinfo{person}{Meirui Jiang}, \bibinfo{person}{Xiaofei Zhang}, {et~al\mbox{.}}} \bibinfo{year}{2021}\natexlab{b}.
\newblock \showarticletitle{FedBN: Federated Learning on Non-IID Features via Local Batch Normalization}. In \bibinfo{booktitle}{\emph{ICLR}}.
\newblock


\bibitem[Liang et~al\mbox{.}(2020)]%
        {liang2020think}
\bibfield{author}{\bibinfo{person}{Paul~Pu Liang}, \bibinfo{person}{Terrance Liu}, \bibinfo{person}{Liu Ziyin}, {et~al\mbox{.}}} \bibinfo{year}{2020}\natexlab{}.
\newblock \showarticletitle{Think Locally, Act Globally: Federated Learning with Local and Global Representations}.
\newblock \bibinfo{journal}{\emph{arXiv:2001.01523}} (\bibinfo{year}{2020}).
\newblock


\bibitem[Lin et~al\mbox{.}(2020)]%
        {lin2020ensemble}
\bibfield{author}{\bibinfo{person}{Tao Lin}, \bibinfo{person}{Lingjing Kong}, \bibinfo{person}{Sebastian~U Stich}, {et~al\mbox{.}}} \bibinfo{year}{2020}\natexlab{}.
\newblock \showarticletitle{Ensemble Distillation for Robust Model Fusion in Federated Learning}. In \bibinfo{booktitle}{\emph{NeurIPS}}. \bibinfo{pages}{2351--2363}.
\newblock


\bibitem[Mansour et~al\mbox{.}(2020)]%
        {mansour2020three}
\bibfield{author}{\bibinfo{person}{Yishay Mansour}, \bibinfo{person}{Mehryar Mohri}, \bibinfo{person}{Jae Ro}, {et~al\mbox{.}}} \bibinfo{year}{2020}\natexlab{}.
\newblock \showarticletitle{Three Approaches for Personalization with Applications to Federated Learning}.
\newblock \bibinfo{journal}{\emph{arXiv:2002.10619}} (\bibinfo{year}{2020}).
\newblock


\bibitem[McMahan et~al\mbox{.}(2017)]%
        {mcmahan2017communication}
\bibfield{author}{\bibinfo{person}{Brendan McMahan}, \bibinfo{person}{Eider Moore}, \bibinfo{person}{Daniel Ramage}, {et~al\mbox{.}}} \bibinfo{year}{2017}\natexlab{}.
\newblock \showarticletitle{Communication-Efficient Learning of Deep Networks from Decentralized Data}. In \bibinfo{booktitle}{\emph{AISTATS}}. \bibinfo{pages}{1273--1282}.
\newblock


\bibitem[Mora et~al\mbox{.}(2022)]%
        {mora2022knowledge}
\bibfield{author}{\bibinfo{person}{Alessio Mora}, \bibinfo{person}{Irene Tenison}, \bibinfo{person}{Paolo Bellavista}, {et~al\mbox{.}}} \bibinfo{year}{2022}\natexlab{}.
\newblock \showarticletitle{Knowledge Distillation for Federated Learning: A Practical Guide}.
\newblock \bibinfo{journal}{\emph{arXiv:2211.04742}} (\bibinfo{year}{2022}).
\newblock


\bibitem[Netzer et~al\mbox{.}(2011)]%
        {netzer2011reading}
\bibfield{author}{\bibinfo{person}{Yuval Netzer}, \bibinfo{person}{Tao Wang}, \bibinfo{person}{Adam Coates}, {et~al\mbox{.}}} \bibinfo{year}{2011}\natexlab{}.
\newblock \showarticletitle{Reading Digits in Natural Images with Unsupervised Feature Learning}. In \bibinfo{booktitle}{\emph{NeurIPS Workshop}}.
\newblock


\bibitem[Ni et~al\mbox{.}(2022)]%
        {ni2022esa}
\bibfield{author}{\bibinfo{person}{Xuanming Ni}, \bibinfo{person}{Xinyuan Shen}, {and} \bibinfo{person}{Huimin Zhao}.} \bibinfo{year}{2022}\natexlab{}.
\newblock \showarticletitle{Federated Optimization via Knowledge Codistillation}.
\newblock \bibinfo{journal}{\emph{Expert Systems with Applications}}  \bibinfo{volume}{191} (\bibinfo{year}{2022}), \bibinfo{pages}{116310}.
\newblock


\bibitem[Oh et~al\mbox{.}(2022)]%
        {oh2022fedbabu}
\bibfield{author}{\bibinfo{person}{Jaehoon Oh}, \bibinfo{person}{Sangmook Kim}, {and} \bibinfo{person}{Seyoung Yun}.} \bibinfo{year}{2022}\natexlab{}.
\newblock \showarticletitle{FedBABU: Toward Enhanced Representation for Federated Image Classification}. In \bibinfo{booktitle}{\emph{ICLR}}.
\newblock


\bibitem[Ozkara et~al\mbox{.}(2021)]%
        {ozkara2021quped}
\bibfield{author}{\bibinfo{person}{Kaan Ozkara}, \bibinfo{person}{Navjot Singh}, \bibinfo{person}{Deepesh Data}, {et~al\mbox{.}}} \bibinfo{year}{2021}\natexlab{}.
\newblock \showarticletitle{QuPeD: Quantized Personalization via Distillation with Applications to Federated Learning}. In \bibinfo{booktitle}{\emph{NeurIPS}}. \bibinfo{pages}{3622--3634}.
\newblock


\bibitem[Peng et~al\mbox{.}(2019)]%
        {peng2019moment}
\bibfield{author}{\bibinfo{person}{Xingchao Peng}, \bibinfo{person}{Qinxun Bai}, \bibinfo{person}{Xide Xia}, {et~al\mbox{.}}} \bibinfo{year}{2019}\natexlab{}.
\newblock \showarticletitle{Moment Matching for Multi-Source Domain Adaptation}. In \bibinfo{booktitle}{\emph{ICCV}}. \bibinfo{pages}{1406--1415}.
\newblock


\bibitem[Pouyanfar et~al\mbox{.}(2018)]%
        {pouyanfar2018survey}
\bibfield{author}{\bibinfo{person}{Samira Pouyanfar}, \bibinfo{person}{Saad Sadiq}, \bibinfo{person}{Yilin Yan}, {et~al\mbox{.}}} \bibinfo{year}{2018}\natexlab{}.
\newblock \showarticletitle{A Survey on Deep Learning: Algorithms, Techniques, and Applications}.
\newblock \bibinfo{journal}{\emph{ACM CSUR}} \bibinfo{volume}{51}, \bibinfo{number}{5} (\bibinfo{year}{2018}), \bibinfo{pages}{1--36}.
\newblock


\bibitem[Qi et~al\mbox{.}(2022)]%
        {qi2022feeling}
\bibfield{author}{\bibinfo{person}{Fan Qi}, \bibinfo{person}{Zixin Zhang}, \bibinfo{person}{Xianshan Yang}, {et~al\mbox{.}}} \bibinfo{year}{2022}\natexlab{}.
\newblock \showarticletitle{Feeling Without Sharing: A Federated Video Emotion Recognition Framework Via Privacy-Agnostic Hybrid Aggregation}. In \bibinfo{booktitle}{\emph{ACM MM}}. \bibinfo{pages}{151--160}.
\newblock


\bibitem[Setayesh et~al\mbox{.}(2023)]%
        {setayesh2023perfedmask}
\bibfield{author}{\bibinfo{person}{Mehdi Setayesh}, \bibinfo{person}{Xiaoxiao Li}, {and} \bibinfo{person}{Vincent~WS Wong}.} \bibinfo{year}{2023}\natexlab{}.
\newblock \showarticletitle{PerFedMask: Personalized Federated Learning with Optimized Masking Vectors}. In \bibinfo{booktitle}{\emph{ICLR}}.
\newblock


\bibitem[Sui et~al\mbox{.}(2020)]%
        {sui2020emnlp}
\bibfield{author}{\bibinfo{person}{Dianbo Sui}, \bibinfo{person}{Yubo Chen}, \bibinfo{person}{Jun Zhao}, {et~al\mbox{.}}} \bibinfo{year}{2020}\natexlab{}.
\newblock \showarticletitle{FedED: Federated Learning via Ensemble Distillation for Medical Relation Extraction}. In \bibinfo{booktitle}{\emph{EMNLP}}. \bibinfo{pages}{2118--2128}.
\newblock


\bibitem[Tan et~al\mbox{.}(2022)]%
        {tan2022towards}
\bibfield{author}{\bibinfo{person}{Alysa~Ziying Tan}, \bibinfo{person}{Han Yu}, \bibinfo{person}{Lizhen Cui}, {et~al\mbox{.}}} \bibinfo{year}{2022}\natexlab{}.
\newblock \showarticletitle{Towards Personalized Federated Learning}.
\newblock \bibinfo{journal}{\emph{IEEE TNNLS}} (\bibinfo{year}{2022}), \bibinfo{pages}{1--17}.
\newblock


\bibitem[Usmanova et~al\mbox{.}(2021)]%
        {usmanova2021distillation}
\bibfield{author}{\bibinfo{person}{Anastasiia Usmanova}, \bibinfo{person}{Fran{\c{c}}ois Portet}, \bibinfo{person}{Philippe Lalanda}, {et~al\mbox{.}}} \bibinfo{year}{2021}\natexlab{}.
\newblock \showarticletitle{A Distillation-based Approach Integrating Continual Learning and Federated Learning for Pervasive Services}. In \bibinfo{booktitle}{\emph{arXiv:2109.04197}}.
\newblock


\bibitem[Wang et~al\mbox{.}(2023)]%
        {wang2023dafkd}
\bibfield{author}{\bibinfo{person}{Haozhao Wang}, \bibinfo{person}{Yichen Li}, \bibinfo{person}{Wenchao Xu}, {et~al\mbox{.}}} \bibinfo{year}{2023}\natexlab{}.
\newblock \showarticletitle{DaFKD: Domain-aware Federated Knowledge Distillation}. In \bibinfo{booktitle}{\emph{CVPR}}. \bibinfo{pages}{20412--20421}.
\newblock


\bibitem[Wen et~al\mbox{.}(2023)]%
        {wen2023communication}
\bibfield{author}{\bibinfo{person}{Hui Wen}, \bibinfo{person}{Yue Wu}, \bibinfo{person}{Jia Hu}, {et~al\mbox{.}}} \bibinfo{year}{2023}\natexlab{}.
\newblock \showarticletitle{Communication-Efficient Federated Learning on Non-IID Data using Two-Step Knowledge Distillation}.
\newblock \bibinfo{journal}{\emph{IEEE IoT}} \bibinfo{volume}{10}, \bibinfo{number}{19} (\bibinfo{year}{2023}), \bibinfo{pages}{17307--17322}.
\newblock


\bibitem[Wen et~al\mbox{.}(2022)]%
        {wen2022communication}
\bibfield{author}{\bibinfo{person}{Hui Wen}, \bibinfo{person}{Yue Wu}, \bibinfo{person}{Jingjing Li}, {et~al\mbox{.}}} \bibinfo{year}{2022}\natexlab{}.
\newblock \showarticletitle{Communication-Efficient Federated Data Augmentation on Non-IID Data}. In \bibinfo{booktitle}{\emph{CVPR}}. \bibinfo{pages}{3377--3386}.
\newblock


\bibitem[Wu et~al\mbox{.}(2022)]%
        {wu2022communication}
\bibfield{author}{\bibinfo{person}{Chuhan Wu}, \bibinfo{person}{Fangzhao Wu}, \bibinfo{person}{Lingjuan Lyu}, {et~al\mbox{.}}} \bibinfo{year}{2022}\natexlab{}.
\newblock \showarticletitle{Communication-Efficient Federated Learning via Knowledge Distillation}.
\newblock \bibinfo{journal}{\emph{Nature Communications}}  \bibinfo{volume}{13} (\bibinfo{year}{2022}), \bibinfo{pages}{2032--2040}.
\newblock


\bibitem[Xu et~al\mbox{.}(2023)]%
        {xu2023personalized}
\bibfield{author}{\bibinfo{person}{Jian Xu}, \bibinfo{person}{Xinyi Tong}, {and} \bibinfo{person}{Shaolun Huang}.} \bibinfo{year}{2023}\natexlab{}.
\newblock \showarticletitle{Personalized Federated Learning with Feature Alignment and Classifier Collaboration}. In \bibinfo{booktitle}{\emph{ICLR}}.
\newblock


\bibitem[Yang et~al\mbox{.}(2023)]%
        {yang2023personalized}
\bibfield{author}{\bibinfo{person}{Lei Yang}, \bibinfo{person}{Jiaming Huang}, \bibinfo{person}{Wanyu Lin}, {et~al\mbox{.}}} \bibinfo{year}{2023}\natexlab{}.
\newblock \showarticletitle{Personalized Federated Learning on Non-IID Data via Group-based Meta-learning}.
\newblock \bibinfo{journal}{\emph{ACM TKDD}} \bibinfo{volume}{17}, \bibinfo{number}{4} (\bibinfo{year}{2023}), \bibinfo{pages}{1--20}.
\newblock


\bibitem[Ye et~al\mbox{.}(2023)]%
        {ye2023pfedsa}
\bibfield{author}{\bibinfo{person}{Chuyao Ye}, \bibinfo{person}{Hao Zheng}, \bibinfo{person}{Zhigang Hu}, {et~al\mbox{.}}} \bibinfo{year}{2023}\natexlab{}.
\newblock \showarticletitle{PFedSA: Personalized Federated Multi-Task Learning via Similarity Awareness}. In \bibinfo{booktitle}{\emph{IPDPS}}. \bibinfo{pages}{480--488}.
\newblock


\bibitem[Zhang et~al\mbox{.}(2023a)]%
        {zhang2023fedala}
\bibfield{author}{\bibinfo{person}{Jianqing Zhang}, \bibinfo{person}{Yang Hua}, \bibinfo{person}{Hao Wang}, {et~al\mbox{.}}} \bibinfo{year}{2023}\natexlab{a}.
\newblock \showarticletitle{FedALA: Adaptive Local Aggregation for Personalized Federated Learning}. In \bibinfo{booktitle}{\emph{AAAI}}. \bibinfo{pages}{11237--11244}.
\newblock


\bibitem[Zhang et~al\mbox{.}(2022)]%
        {zhang2022fine}
\bibfield{author}{\bibinfo{person}{Lin Zhang}, \bibinfo{person}{Li Shen}, \bibinfo{person}{Liang Ding}, {et~al\mbox{.}}} \bibinfo{year}{2022}\natexlab{}.
\newblock \showarticletitle{Fine-tuning Global Model via Data-Free Knowledge Distillation for Non-IID Federated Learning}. In \bibinfo{booktitle}{\emph{CVPR}}. \bibinfo{pages}{10174--10183}.
\newblock


\bibitem[Zhang et~al\mbox{.}(2021)]%
        {zhang2020personalized}
\bibfield{author}{\bibinfo{person}{Michael Zhang}, \bibinfo{person}{Karan Sapra}, \bibinfo{person}{Sanja Fidler}, {et~al\mbox{.}}} \bibinfo{year}{2021}\natexlab{}.
\newblock \showarticletitle{Personalized Federated Learning with First Order Model Optimization}. In \bibinfo{booktitle}{\emph{ICLR}}.
\newblock


\bibitem[Zhang et~al\mbox{.}(2023b)]%
        {zhang2023fedmpt}
\bibfield{author}{\bibinfo{person}{Xinglin Zhang}, \bibinfo{person}{Zhaojing Ou}, {and} \bibinfo{person}{Zheng Yang}.} \bibinfo{year}{2023}\natexlab{b}.
\newblock \showarticletitle{FedMPT: Federated Learning for Multiple Personalized Tasks Over Mobile Computing}.
\newblock \bibinfo{journal}{\emph{IEEE TNSE}} \bibinfo{volume}{10}, \bibinfo{number}{4} (\bibinfo{year}{2023}), \bibinfo{pages}{2358--2371}.
\newblock


\bibitem[Zheng et~al\mbox{.}(2023)]%
        {zheng2023federated}
\bibfield{author}{\bibinfo{person}{Xiaolin Zheng}, \bibinfo{person}{Senci Ying}, \bibinfo{person}{Fei Zheng}, {et~al\mbox{.}}} \bibinfo{year}{2023}\natexlab{}.
\newblock \showarticletitle{Federated Learning on Non-IID Data via Local and Global Distillation}.
\newblock \bibinfo{journal}{\emph{arXiv:2306.14443}} (\bibinfo{year}{2023}).
\newblock


\bibitem[Zhu et~al\mbox{.}(2021)]%
        {zhu2021data}
\bibfield{author}{\bibinfo{person}{Zhuangdi Zhu}, \bibinfo{person}{Junyuan Hong}, {and} \bibinfo{person}{Jiayu Zhou}.} \bibinfo{year}{2021}\natexlab{}.
\newblock \showarticletitle{Data-Free Knowledge Distillation for Heterogeneous Federated Learning}. In \bibinfo{booktitle}{\emph{ICML}}. \bibinfo{pages}{12878--12889}.
\newblock


\bibitem[Zhuang et~al\mbox{.}(2020)]%
        {zhuang2020performance}
\bibfield{author}{\bibinfo{person}{Weiming Zhuang}, \bibinfo{person}{Yonggang Wen}, \bibinfo{person}{Xuesen Zhang}, {et~al\mbox{.}}} \bibinfo{year}{2020}\natexlab{}.
\newblock \showarticletitle{Performance Optimization of Federated Person Re-identification via Benchmark Analysis}. In \bibinfo{booktitle}{\emph{ACM MM}}. \bibinfo{pages}{955--963}.
\newblock


\end{thebibliography}

\end{document}